%% file: arxiv.tex
\documentclass{article}

\usepackage[preprint]{neurips_2026}

\input{neurips_2026/neurips_packages}

\usepackage[most]{tcolorbox}

\title{Diffusion Model Attribution via Spectral Coupling of Denoiser Responses}
\author{%
  Pragati Shuddhodhan Meshram \quad Varun Chandrasekaran\\
  Department of Electrical and Computer Engineering\\
  University of Illinois Urbana-Champaign\\
  \texttt{\{psm12,varunc\}@illinois.edu}
}

\begin{document}

\maketitle

\input{neurips_2026/arxiv_section/0_abstract_arxiv}

\input{neurips_2026/arxiv_section/1_introduction_arxiv}
\input{neurips_2026/arxiv_section/2_related_work_arxiv}
\input{neurips_2026/arxiv_section/3_motivation_arxiv}
\input{neurips_2026/arxiv_section/4_method_arxiv}

\input{neurips_2026/arxiv_section/6_experimental_setup_arxiv}
\input{neurips_2026/arxiv_section/7_results_arxiv}

\input{neurips_2026/arxiv_section/9_conclusion_arxiv}

\bibliography{neurips_2026/neurips_2026}
\bibliographystyle{plainnat}

\newpage
\input{neurips_2026/arxiv_section/appendix_arxiv}
\end{document}

%% file: neurips_2026/neurips_packages.tex
\usepackage[utf8]{inputenc} 
\usepackage[T1]{fontenc}    
\usepackage{hyperref}       
\usepackage{url}            
\usepackage{graphicx}
\usepackage{amssymb}
\usepackage{booktabs}       
\usepackage{amsfonts}       
\usepackage{nicefrac}       
\usepackage{microtype}      
\usepackage{xcolor}         
\usepackage{enumitem}
\usepackage[ruled,vlined]{algorithm2e}
\RestyleAlgo{boxed}
\usepackage{wrapfig}
\usepackage{svg}
\usepackage{amsmath}

  \newenvironment{squishenumerate}
  {\begin{list}{\arabic{enumi}.}{%
    \usecounter{enumi}%
    \setlength{\itemsep}{0pt}%
    \setlength{\parsep}{0pt}%
    \setlength{\topsep}{0pt}%
    \setlength{\parskip}{0pt}%
    \setlength{\labelwidth}{.5in}%
    \setlength{\labelsep}{0.05in}%
    \setlength{\leftmargin}{.2in}}}
  {\end{list}}

\usepackage{xcolor}

\usepackage{xcolor}

%% file: neurips_2026/arxiv_section/0_abstract_arxiv.tex
\begin{abstract}
Attributing a generated image to its source diffusion model is a fundamental challenge in provenance verification and intellectual property protection. This problem is particularly difficult because diffusion models trained on different datasets can converge to similar score functions and thus similar output distributions, making the generated images themselves unreliable as attribution evidence.
Existing non-invasive methods either fail on architecturally similar variants or rely on signals that vanish when models share the same autoencoder. We propose \textit{Spectral Denoising Signatures} (\textsc{SDS}), a non-invasive attribution method that identifies the source model by fingerprinting each candidate model's denoising behavior. Our key insight is that a model’s denoising score function exhibits a distinctive spectral geometry, reflected in how it redistributes energy across spatial frequency bands during denoising. By probing this behavior with frequency-controlled perturbations, \textsc{SDS} extracts a stable signature that is intrinsic to the model, requiring only standard forward passes with no inversion, optimization, or generation-time enrollment. Our results demonstrate that \textsc{SDS} achieves $ \approx 99.9\%$ 
accuracy across eight diverse diffusion models and $96.2\%$ under cross-domain prompt shift, outperforming the non-invasive baselines across variations in training data, architecture, and training procedure, establishing spectral geometry as a principled and practical basis for diffusion model attribution. 
Our code is available at \url{https://github.com/Pragati-Meshram/SGS}.
\end{abstract}

%% file: neurips_2026/arxiv_section/1_introduction_arxiv.tex
\section{Introduction}
\label{sec:intro}


Diffusion models have become the dominant paradigm for 
large-scale generative modeling, with systems such as 
Stable Diffusion~\citep{rombach2022high}, 
SDXL~\citep{podell2023sdxl}, and 
PixArt-$\alpha$~\citep{chen2023pixart} enabling 
high-fidelity image synthesis at unprecedented scale. As 
these models are widely distributed and fine-tuned in open 
ecosystems, the need to verify model provenance has become 
increasingly critical for ownership verification, 
provenance tracking, and misuse detection. In particular, 
\emph{given a generated image, can we reliably determine 
which model produced it?}



This problem is fundamentally challenging. Diffusion 
models generate images through a stochastic, multi-step 
denoising process, in which noise is gradually transformed 
into a final image via a learned score function, rather 
than through a single forward mapping. Prior work has 
shown that even models trained on different datasets can 
converge to remarkably similar score functions, producing 
visually indistinguishable outputs~\citep{kadkhodaie2023generalization}. 
Attribution methods that rely solely on generated images 
therefore face an inherent ambiguity, with limited 
discriminative signal in the output space.


Existing approaches to model attribution fall into
two broad families, each with fundamental
limitations. \emph{Invasive methods} such as
watermarking require retraining or fine-tuning and
may degrade generation
quality~\citep{zhao2023recipe, fernandez2023stable,
yuan2024watermarking}. \emph{Non-invasive methods}
avoid model modification but remain limited in what
they can distinguish. Among these, classifier-based
approaches~\citep{xu2025detecting, corvi2023intriguing}
operate purely on generated images and struggle to
separate architecturally similar variants that produce
visually similar outputs. Inversion-based methods such
as \textsc{RONAN}~\citep{wang2023did} and
\textsc{LatentTracer}~\citep{wang2024trace} go further by
querying model internals, but require iterative
optimization per candidate and lose discriminative
power when candidates share the same autoencoder:
reconstruction losses become uninformative regardless
of which model is queried~\citep{wang2024trace}.
The fundamental gap is the same across all 
non-invasive approaches: no existing method 
probes the denoising function itself without 
costly inversion or reliance on output-space 
signals that vanish for similar models.

A complementary line of work targets model ownership 
verification rather than source attribution. 
\citet{teng2025fingerprinting} introduced 
\textsc{FingerInv},
the first non-invasive fingerprinting 
method for diffusion models, establishing the important 
principle 
that model-specific signal resides 
in the denoiser, not in generated outputs.
However, its signature is highly tied to a specific chosen image rather than being intrinsic to the model, limiting 
generalization to arbitrary generated images.



\begin{figure}
    \centering
    \includegraphics[width=0.85\linewidth]{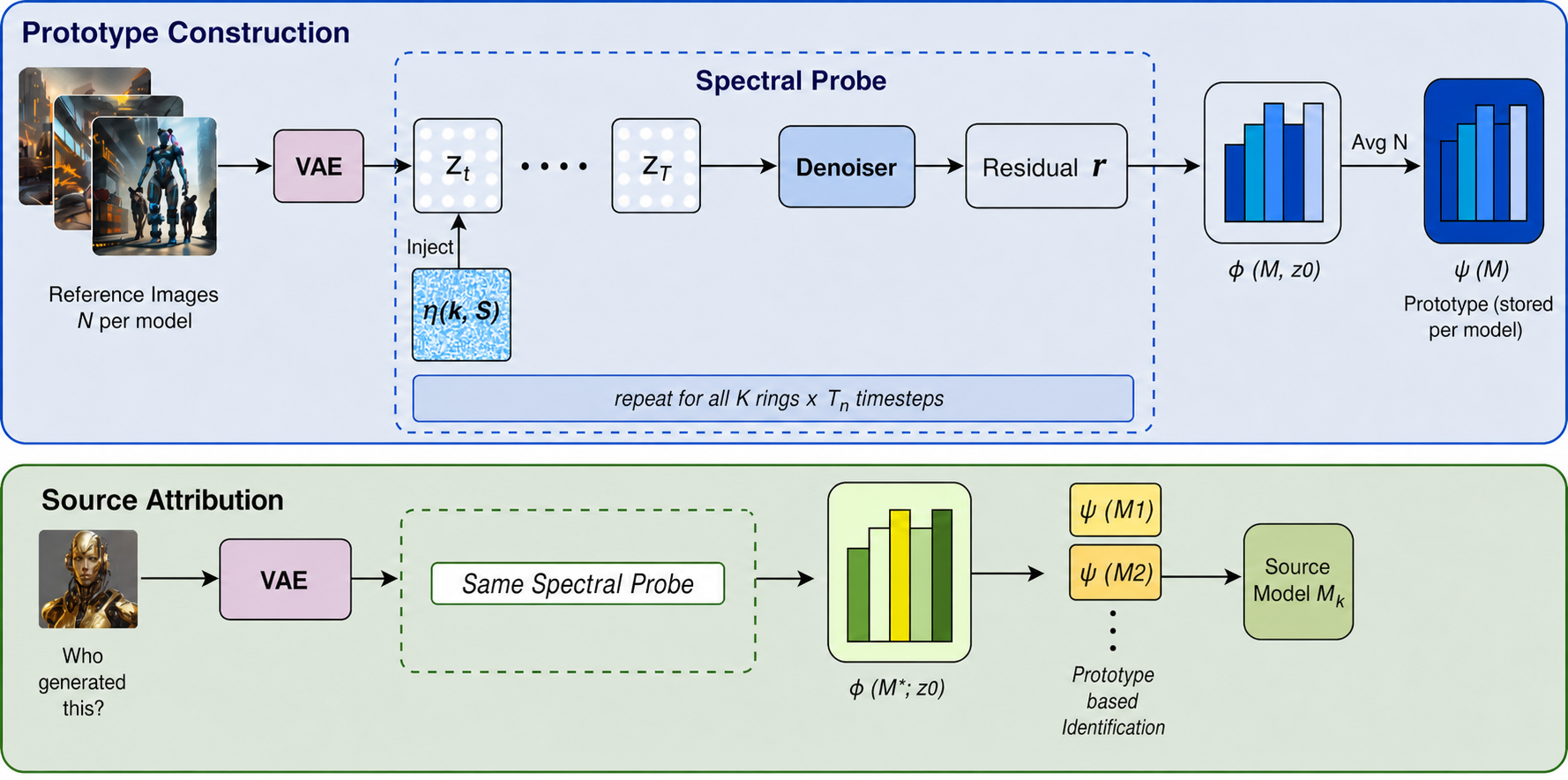}
    \caption{\footnotesize \textbf{Pipeline for Spectral Denoising Signatures (\textsc{SDS})}. Prototype Construction (top): for each candidate model $M$, reference images are encoded via a shared VAE to latent $z_0$. The spectral probe injects band-limited noise $\eta(k, S)$ into frequency ring $k$, runs a single denoiser forward pass, and measures cross-band energy in the residual $\mathbf{r}$, forming coupling matrix. Repeating over all $K$ rings and $T_n$ timesteps yields signature $\phi(M, z_0)$; averaging over $N$ such signatures produces prototype $\psi(M)$. Source Attribution (bottom): a query image of unknown origin is encoded and probed identically. The resulting signature $\phi(M^*, z_0)$ is compared against stored prototypes using classification techniques; the source model $M_k$ is identified as the nearest prototype.\vspace{-9pt}}
    \label{fig:pipeline}
\end{figure}
In this work, we take a different perspective. 
Instead of attributing images by analyzing \emph{what 
a model generates}, we attribute them by 
characterizing \emph{how a model denoises}. To this end, we propose \textbf{Spectral Denoising Signatures} (\textsc{SDS}), a non-invasive model attribution method that characterizes 
diffusion models through the spectral structure of their 
score function (see Fig.~\ref{fig:pipeline}). Our key insight is that each model exhibits a distinctive \emph{spectral geometry}, reflected in how it redistributes energy across spatial frequency bands during the denoising process. By injecting frequency-controlled perturbations and measuring cross-band energy transfer over timesteps, \textsc{SDS} extracts a stable signature that is intrinsic to the model. The method requires only standard forward passes, with no inversion, optimization, or model modification.

Beyond proposing a new attribution method, we also aim to understand \emph{what factors give rise to model identity}. We evaluate \textsc{SDS} on a challenging closed-set task spanning eight diffusion models that differ in 
training data, architecture, and training procedure. This enables us to answer key questions left open by prior work: can models with identical architectures but different datasets be distinguished? Can fingerprints persist across different architectures trained on similar data? Does distillation leave a detectable imprint on an otherwise identical model? Our results show that \textsc{SDS} reliably captures all three sources of variation, providing a unified view of how diffusion models encode their identity.

In summary, we make three contributions: (1) we 
propose \textsc{SDS}, a non-invasive attribution method 
that probes the denoiser's spectral geometry via 
forward pass alone, requiring no inversion, 
optimization, or model modification; (2) 
we demonstrate that spectral coupling signatures 
are model-intrinsic, distortion-robust, and 
discriminative across all sources of model 
variation including architecture, training data, and training procedure, with only a $3.8$pp 
accuracy drop under cross-domain prompt shift; and (3) we show that 
probing the denoiser rather than the decoder 
resolves the fundamental failure mode of existing 
inversion-based baselines.

\vspace{-13pt}

%% file: neurips_2026/arxiv_section/2_related_work_arxiv.tex
\section{Background \& Related Work}
\vspace{-5pt}

\noindent{\bf Diffusion denoising.}
Latent diffusion models~\citep{rombach2022high} 
define a forward process that gradually corrupts 
a latent $z_0$ with Gaussian noise across $T$ 
timesteps~\citep{ho2020denoising}, producing 
$z_t = \sqrt{\bar{\alpha}_t}\,z_0 + 
\sqrt{1{-}\bar{\alpha}_t}\,\eta$ where 
$\eta \sim \mathcal{N}(0,I)$. A denoising 
network $\hat{\epsilon}_\theta(z_t, t)$ is trained 
to reverse this process by estimating the noise 
component at each noise level, which is equivalent 
to approximating the score function 
$\nabla_{z_t} \log p_t(z_t)$ of the data 
distribution~\citep{song2020score}. At inference, 
a sampler such as DDIM~\citep{song2020denoising} 
iteratively applies the denoiser from $t{=}T$ 
down to $t{=}0$ to produce a clean image from 
pure noise. Because this network parameterizes
the score function of the training distribution,
any model-specific inductive bias must manifest
through it, making it the right object to probe
for attribution.

\noindent{\bf Fingerprinting diffusion models.}
Diffusion models trained on non-overlapping 
subsets of a dataset converge to nearly identical 
score functions~\citep{kadkhodaie2023generalization}, 
suggesting that output-based attribution is 
fundamentally limited when training distributions 
are similar.
\citet{teng2025fingerprinting} propose
\textsc{FingerInv} for model ownership
verification, which identifies models via inversion trajectories of generated images. Other approaches embed fingerprints directly into models through weight modulation or decoder fine-tuning for user attribution~\citep{tripathi2025paladin, fei2025omnimark}, but these require modifying model parameters and training pipelines. In contrast, we show that despite apparent convergence, diffusion models exhibit distinct spectral dynamics during denoising, enabling intrinsic and non-invasive fingerprinting.

\noindent{\bf Non-invasive model attribution.}
A growing line of work addresses source attribution 
without modifying the model or its outputs. 
Classifier-based methods~\citep{xu2025detecting, 
bonechi2025made} train on output-space artifacts to 
identify source generators, achieving strong 
performance across diverse models but struggling to 
separate architecturally similar variants that produce 
visually similar outputs. Inversion-based detectors such as \textsc{RONAN}~\citep{wang2023did} and \textsc{LatentTracer}~\citep{wang2024trace} attribute an image to the model that reconstructs it with the smallest calibrated loss; they are alteration-free yet slow (one optimisation per candidate) and lose discriminative power when models share the same auto-encoder.  In contrast, 
\textsc{SDS} probes the denoising function directly, succeeding 
where decoder-level and output-level signals collapse.

In summary, 
existing non-invasive methods exploit output-space
or decoder-level signals that collapse for
shared-autoencoder models. Ownership verification
methods probe the denoiser but require prior
enrollment and solve a different problem.
No existing method probes the denoising function
directly without costly inversion or prior
enrollment. This gap motivates \textsc{SDS}, introduced
after establishing the empirical basis for
denoiser-level probing in the next section.
Additional background on watermarking methods 
for diffusion models and model 
fingerprinting in neural networks is 
provided in Appendix~\ref{app:related}.
\vspace{-13pt}

%% file: neurips_2026/arxiv_section/3_motivation_arxiv.tex
\section{Where Does Model Identity Reside? 
}
\label{sec:motivation}
\vspace{-5pt}
We identify a deliberately hard attribution 
setting to probe where discriminative signal 
resides: \textbf{SD\,v1.4\footnote{\url{https://huggingface.co/CompVis/stable-diffusion-v1-4}}  vs.\ SD\,v1.5\footnote{\url{https://huggingface.co/runwayml/stable-diffusion-v1-5}}}, two 
models sharing nearly all of their weights, 
the same UNet-860M backbone, encoder, and KL-f8 
VAE ~\citep{rombach2022high}. We ask: \emph{where does discriminative 
signal actually reside, if not in outputs?}

\begin{figure}
    \centering
    \includegraphics[width=0.98\linewidth]{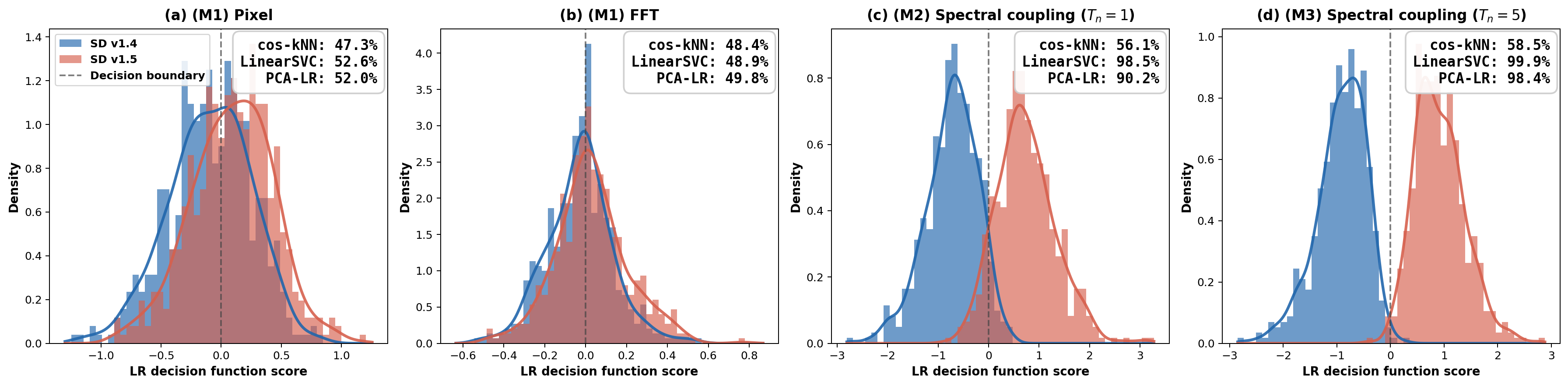}
    \caption{\footnotesize \textbf{Spectral coupling signatures are 
    linearly separable; image features are not.} LR decision score 
    distributions for SD\,v1.4 (blue) vs.\ SD\,v1.5 (red) after 
    PCA-64\,+\,L$_2$ normalisation. Pixel and FFT scores overlap 
    completely, confirming the absence of any linear signal. Spectral 
    Coupling $T_n = 1$ and $T_n = 5$ show clearly separated 
    distributions, directly visualising the global mean-shift that 
    enables $\geq 98\%$ classification accuracy from a 
    192-dimensional signature.}
    \label{fig:motiv_plot}
    \vspace{-20pt}
\end{figure}

\noindent{\bf The denoiser is the right object of study.} 
The answer lies in the denoising function itself ~\citep{teng2025fingerprinting}. 
The denoiser parameterizes the score function 
$\nabla_z \log p(z)$, which encodes the full 
statistical geometry of the training 
distribution, including its cross-frequency 
covariance structure. Any model-specific inductive 
bias arising from training data, architecture, or 
fine-tuning must manifest through this function. 
Unlike output images, which are mediated by specific 
prompts and noise seeds, the denoiser is an 
\emph{intrinsic, input-independent object}: it is 
shared across all generations and directly reflects 
what the model has learned. This motivates us to 
probe the denoiser directly, rather than analyzing 
its outputs.

\noindent{\bf Setup.}
We generate $N{=}1{,}000$ images (500 per model) using 500 prompts from
the Stable Diffusion Prompts
dataset,\footnote{\url{https://huggingface.co/datasets/Gustavosta/Stable-Diffusion-Prompts}}
at $512{\times}512$ resolution with 50 DDIM steps and classifier-free guidance (CFG) scale $7.5$. 
We compare three feature families: pixel-space 
statistics, FFT power spectrum, and spectral 
coupling signatures derived from denoiser 
responses. For fair comparison, all features 
are preprocessed identically: standardization, 
PCA to 64 dimensions, and $\ell_2$ 
normalization. We evaluate three classifiers: 
LinearSVC on raw features, PCA-LR on compressed 
features, and cosine-$k$NN. Results are 
in Figure~\ref{fig:motiv_plot}, 
which shows LR decision score distributions 
for all four feature types.

\textbf{$\text{P}_1$: Output features carry no signal.}
We evaluate pixel-space features ($786,000$ dimensions) and FFT power
spectrum features ($262,000$ dimensions) as standard output-based
representations.
Despite their dimensional advantage, all three classifiers achieve
near-chance performance.
The failure is structural: SD\,v1.4 and SD\,v1.5 share the same VAE,
so their output distributions are statistically indistinguishable
at the level of spatial and spectral statistics that these features
capture.

\noindent \underline{\bf Takeaway 1.}
For models sharing the same autoencoder, output-space features provide
negligible discriminative signal regardless of (high) dimensionality.
This structural failure defines the hard regime where existing
non-invasive methods collapse. 

\textbf{$\text{P}_2$: Model identity emerges in denoiser responses.}
We probe the denoiser's response to controlled frequency perturbations.
For each spatial frequency band $k$, we inject band-limited noise
at a single probe timestep ($T_n{=}1$, 
$t{=}25$, where $T_n$ denotes the number of 
timesteps used for probing, distinct from the 
total diffusion steps $T{=}50$),
chosen to sit
in the mid-noise regime where 
the denoiser Jacobian 
$J_\theta = \partial\hat{\epsilon}_\theta / 
\partial z_t$ (formalized in 
\S\ref{subsec:theory}) captures both global 
and local structure
and measure how the injected energy is redistributed
across output bands.
This yields a 
spectral coupling matrix $C \in 
\mathbb{R}^{K \times K}$, where $K$ is the 
number of frequency rings (a design parameter, 
set to $K{=}8$ here) and $C[k,j]$ quantifies 
how energy injected in ring $k$ is routed to 
ring $j$ by the denoiser.
We probe at randomly chosen $S{=}3$ perturbation amplitudes 
$\{1.0,\,1.25,\,1.5\}$, yielding a 
192-dimensional signature 
($S{\cdot}K^2 = 3{\cdot}64 = 192$).
Detailed method is provided in \S~\ref{subsec:approach}.


LinearSVC achieves $\mathbf{98.5\%}$ accuracy on this compact 192-dimensional representation, compared to near-chance performance for image-based features with over $10^5$ dimensions. PCA-LR (64-dimensional subspace) reaches $90.2\%$, confirming that the discriminative signal is preserved under strong compression. In contrast, cosine-$k$NN performs substantially worse ($56.1\%$), indicating that separation does not arise from tight local clustering. Instead, the signal manifests as a \emph{global mean shift}: class centroids are well separated, while individual samples overlap locally. As a result, linear classifiers recover the signal, whereas neighborhood-based methods fail to capture it. More broadly, these results show that the discriminative information is not in raw outputs themselves, but in how the denoiser \emph{transforms structured perturbations}.

\noindent \underline{\bf Takeaway 2.}
A single-timestep spectral response linearly separates two
very identical models from a 192-dimensional signature.
The signal arises from the denoiser's Jacobian structure and is
therefore intrinsic to the model, not an artifact of the probing
procedure.

\textbf{$\text{P}_3$: Identity accumulates across timesteps.}
We extend the probe to $T_n{=}5$ evenly spaced timesteps spanning
$[0.2T, T]$, concatenating coupling matrices at each step to yield
a 960-dimensional signature
($S \cdot T_n \cdot K^2 = 3 \cdot 5 \cdot 64 = 960$).
LinearSVC reaches $\mathbf{99.9\%}$ and PCA-LR reaches $98.4\%$,
both improving consistently over the single-timestep result.
The gain reflects genuine complementarity rather than added
dimensionality: at high noise levels, the Jacobian primarily encodes
low-frequency global structure, while at low noise levels it encodes
high-frequency texture~\citep{kadkhodaie2023generalization}.
Timesteps therefore probe orthogonal aspects of the model's spectral
geometry, and their signals accumulate.

\noindent \underline{\bf Takeaway 3.}
Aggregating responses across $T_n{=}5$ 
timesteps improves attribution accuracy, indicating that 
different timesteps contribute complementary 
model-specific information that accumulates 
along the denoising trajectory.


Together, these results show that model identity is encoded in the
spectral geometry of the denoiser's Jacobian, not in generated outputs.
These findings motivate \textsc{SDS}: a method that fingerprints models by
probing spectral coupling at multiple timesteps, requiring only
standard forward passes with no inversion or model modification.
\vspace{-13pt}

%% file: neurips_2026/arxiv_section/4_method_arxiv.tex
\section{Spectral Denoising Signatures}
\label{sec:method}
\vspace{-5pt}
We now formalize the extraction of spectral 
coupling signatures from diffusion model 
denoisers into a complete attribution pipeline, 
translating the empirical observations of 
\S~\ref{sec:motivation} into a principled method.
\vspace{-13pt}
\input{neurips_2026/arxiv_section/4.1_overview_arxiv}
\input{neurips_2026/arxiv_section/4.2_steps_arxiv}

%% file: neurips_2026/arxiv_section/4.1_overview_arxiv.tex
\subsection{Problem Formulation}
\label{subsec:problem}
\vspace{-5pt}
\noindent{\bf Task.}
We consider \emph{closed-set model attribution}: given a generated
image $x_0$ and a finite registry of candidate models
$\mathcal{M} = \{M_1, \ldots, M_n\}$, identify which model in
$\mathcal{M}$ produced $x_0$.
The registry may contain models that differ in training data,
architecture, or training procedures, including near-clone variants
(e.g., fine-tuned derivatives with high weight overlap).

\textbf{\bf Verifier access.}
\textsc{SDS} assumes white-box access to candidate models: 
the verifier queries the denoiser forward pass 
$\hat{\epsilon}_\theta(z_t, t, \varnothing)$ 
with arbitrary inputs, where $\varnothing$ 
denotes the empty text embedding used for 
unconditional denoising,
requiring only forward passes 
with no gradients, optimization, or knowledge of 
generation parameters.
\vspace{-10pt}

%% file: neurips_2026/arxiv_section/4.2_steps_arxiv.tex
\subsection{Our Approach}
\label{subsec:approach}
\vspace{-5pt}
\smallskip \noindent{\bf Preliminaries.}
Before probing, we standardize conditions across 
all candidate models to ensure signatures reflect 
genuine model-specific behavior rather than artifacts of mismatched latent spaces or sampling 
parameters. Each input 
image $x_0$ is encoded to a latent $z_0 = 
\textit{E}(x_0)$ using a single shared VAE, 
regardless of which candidate model is being 
probed, yielding consistent latent dimensions 
across all models. A shared scheduler is used for 
all models, ensuring timestep indices refer to 
the same noise levels. The denoiser is called 
with an empty text embedding $\varnothing$ 
throughout, removing prompt-dependent signal and 
ensuring the signature reflects the model's 
intrinsic denoising geometry rather than its 
response to any particular text.

\noindent \underline{\bf 1) Partition.} We partition the $H \times W$ frequency plane of 
the latent into $K$ concentric radial rings. The 
normalised radial coordinate of frequency $(u,v)$ 
is $\rho(u,v) = \frac{\sqrt{u^2+v^2}}{\sqrt{H^2+W^2}}$,
for $k \in \{0,\ldots,K{-}1\}$, ring $k$ 
contains all frequencies whose normalised 
radius $\rho$ falls in 
$\left[\tfrac{k}{K},\,\tfrac{k+1}{K}\right)$
$M_k \in \{0,1\}^{H\times W}$. Ring $0$ covers 
the DC component and lowest spatial frequencies 
(global structure); ring $K{-}1$ covers the 
highest frequencies (fine texture). 

\noindent \underline{\bf 2) Probe.} As established in \S~\ref{sec:motivation}, 
the spectral coupling matrix $C[k,j]$ is a 
frequency-domain projection of the denoiser 
Jacobian $\partial\hat{\varepsilon}_\theta / 
\partial z_t$, 
directly measuring how the score 
function's local geometry routes energy across 
frequency bands. We now formalize this 
measurement. A formal derivation connecting $C[k,j]$ to 
the denoiser Jacobian is provided in 
Appendix~\ref{subsec:theory}.

\textbf{Band-limited probe noise.}
Given i.i.d.\ Gaussian noise 
$\eta \sim \mathcal{N}(0,I)$,
we confine its energy to ring $k$ by masking in the Fourier 
domain and renormalising to match the original 
noise energy:
{
\setlength{\abovedisplayskip}{1pt}
\setlength{\belowdisplayskip}{1pt}
\begin{equation}
  \eta^{(k)} =
    \mathcal{F}^{-1}\!\bigl(
    \mathcal{F}(\eta)\odot M_k\bigr)
    \cdot \frac{\sigma(\eta)}{\sigma\!\left(
      \mathcal{F}^{-1}(
      \mathcal{F}(\eta)\odot M_k)
    \right) + \delta}
  \label{eq:bandnoise}
\end{equation}}The renormalisation ($\delta = 10^{-8}$) ensures 
all rings are probed at comparable signal levels. 
A scaled version with amplitude $s$ is then
$\eta^{(k,s)} = s \cdot \eta^{(k)}$.

\textbf{Denoiser query.}
We construct the noisy latent and query the 
denoiser:
{
\setlength{\abovedisplayskip}{1pt}
\setlength{\belowdisplayskip}{1pt}
\begin{align}
  z_t &= \sqrt{\bar{\alpha}_t}\,z_0
         + \sqrt{1{-}\bar{\alpha}_t}\,
         \eta^{(k,s)},
  \label{eq:noisy} \\
  \hat{\eta} &= \hat{\epsilon}_\theta(
  z_t,\,t,\,\varnothing)
  \label{eq:denoise}
\end{align}
}The residual $\mathbf{r} = \hat{\eta} - 
\eta^{(k,s)}$ measures how the model's 
predicted noise deviates from the injected 
perturbation. If the model were a perfect noise 
estimator for band $k$, the residual would be 
zero; any nonzero residual reveals which 
frequencies the denoiser maps energy into or 
out of.

\textbf{Coupling matrix entry.}
For a fixed latent $z_0$, the denoiser response 
to a band-limited perturbation is stochastic; 
we therefore average over $R$ independently 
drawn noise samples to obtain a stable estimate 
of ring-$k$ to ring-$j$ energy routing:
{
\setlength{\abovedisplayskip}{1pt}
\setlength{\belowdisplayskip}{1pt}
\begin{equation}
  C_{s,t}[k,j]
  = \frac{1}{R}\sum_{i=1}^{R}
    \bigl\langle \lvert\mathcal{F}(
    \mathbf{r}^{(k,s,t)}_i)\rvert^2,\,M_j
    \bigr\rangle
  \label{eq:coupling}
\end{equation}
}Row $k$ of $C_{s,t}$ encodes the 
denoiser's output energy distribution when 
ring $k$ is perturbed, capturing which 
output frequency bands this model routes 
that energy to

\textbf{Row normalisation.}
Each row is $\ell_1$-normalised so that it 
sums to one: 
{
\setlength{\abovedisplayskip}{1pt}
\setlength{\belowdisplayskip}{1pt}
\begin{equation}
  C_{s,t}[k,\cdot]
  \leftarrow
  \frac{C_{s,t}[k,\cdot]}
       {\|C_{s,t}[k,\cdot]\|_1 + \delta}
  \label{eq:rownorm}
\end{equation}
}This converts absolute energies into a 
distribution over output bands, removing 
content-dependent scale variation so that the 
coupling pattern reflects the model's learned 
behavior rather than image brightness.

\noindent \underline{\bf 3) Aggregate.}
We collect coupling measurements across 
multiple timesteps and perturbation amplitudes 
to build a complete model signature that 
captures spectral routing behavior along the 
full denoising trajectory.

\textbf{Timestep selection.}
We sample $T_n$ evenly spaced timesteps 
in $T$, spanning overall 
denoising behavior as motivated 
by Takeaway~3 in \S~\ref{sec:motivation}. 

\textbf{Multi-scale probing.}
We evaluate the denoiser's response across $S$ 
perturbation amplitudes, chosen to span a range 
of noise magnitudes while remaining within the 
model's operational noise regime. Aggregating 
across scales captures how spectral coupling 
behaves under different perturbation strengths, 
providing complementary discriminative signal 
with minimal computational overhead.

\textbf{Signature tensor and final vector.}
The full measurement is collected into:
{
\setlength{\abovedisplayskip}{0pt}
\setlength{\belowdisplayskip}{1pt}
\begin{equation}
  \mathbf{A} \in \mathbb{R}^{S \times T_n \times 
  K \times K},
  \qquad
  \mathbf{A}[s,\,t,\,k,\,j] = 
  C_{s,t}[k,j]
  \label{eq:tensor}
\end{equation}
}The \textsc{SDS} signature is the flattened tensor:
{
\setlength{\abovedisplayskip}{1pt}
\setlength{\belowdisplayskip}{1pt}
\begin{equation}
  \Phi(M,\,z_0) = \mathrm{vec}(\mathbf{A})
  \;\in\; \mathbb{R}^{S \cdot T_n \cdot K^2}
  \label{eq:signature}
\end{equation}
}Each dimension encodes one specific query: 
\emph{at timestep $t$ and amplitude $s$, if 
energy is injected in ring $k$, what fraction 
of the denoiser residual appears in ring $j$?}

\noindent \underline{\bf 4) Identify.}
Given a query image, we compare its signature 
against stored model prototypes to determine 
the source model, either via nearest-prototype 
matching or a trained linear classifier.
\textbf{Prototype construction.}
Given $N$ reference images $\{x_0^{(i)}\}$ 
generated by model $M$, the model prototype is 
the mean signature:
{
\setlength{\abovedisplayskip}{1pt}
\setlength{\belowdisplayskip}{1pt}
\begin{equation}
  \Psi(M) = \frac{1}{N}\sum_{i=1}^{N}
             \Phi\!\left(M,\,
             \textit{E}(x_0^{(i)})\right)
  \label{eq:proto}
\end{equation}
}Averaging suppresses content-dependent noise, 
recovering the model's intrinsic centroid in 
signature space. The row normalisation in 
Eq.~\ref{eq:rownorm} removes content-scale 
variation by design, which is why a small $N$ 
suffices. 

\textbf{Zero-training identification (argmin).}
Given a test image from an unknown model $M^*$, 
we assign it to the nearest prototype:
{
\setlength{\abovedisplayskip}{1pt}
\setlength{\belowdisplayskip}{1pt}
\begin{equation}
  \hat{M} = \underset{M \in \mathcal{M}}{\arg\min}
  \;d_{\cos}\!\left(\Phi(M^*,\,\textit{E}(x)),
  \;\Psi(M)\right),
  \label{eq:argmin}
\end{equation}
}where $d_{\cos}(u,v)=\frac{1-u^\top v}{(\|u\|\|v\|)}$.
This requires no classifier training; 
identification is immediate once prototypes 
are computed.


\textbf{Classifier-based identification.}
When labelled signatures are available, a 
LinearSVC trained on raw signatures 
substantially improves over argmin, 
particularly when model weight overlap is 
high. That a \emph{linear} classifier 
suffices reflects the global mean-shift 
structure from \S~\ref{sec:motivation}: 
model classes are linearly separable in 
signature space, making nonlinear 
alternatives unnecessary. The full algorithm 
is in Algorithm~\ref{alg:sgs} 
(Appendix~\ref{app:algorithm}).

%% file: neurips_2026/arxiv_section/6_experimental_setup_arxiv.tex
\section{Experimental Setup}
\label{sec:experimental}
\vspace{-5pt}

We design a single closed-set attribution 
experiment spanning eight diffusion models 
representing distinct combinations of training 
data, architecture, and training procedure. 
We additionally evaluate under six strong image-level distortions i.e.,  
rotation ($75^\circ$), JPEG compression 
(Q=25), centre crop ($75\%$), brightness 
scaling (${\times}1.5$), Gaussian blur 
($\sigma{=}8$), and additive noise 
($\sigma{=}0.1$), to assess robustness. 

\textbf{Training data variation (fixed ``1.x'' latent architecture).}
Stable Diffusion v1.4, v1.5, Dreamshaper-8\footnote{\url{https://huggingface.co/Lykon/dreamshaper-8}}, and Realistic Vision v5\footnote{\url{https://huggingface.co/SG161222/Realistic_Vision_V5.1_noVAE}}
all build on the same SD~1.x UNet-860M latent diffusion backbone, but
differ in training data distribution, ranging from base LAION~\citep{schuhmann2022laion} training 
to domain-specific fine-tuning.
Stable Diffusion v2.1 extends this family with updated data curation
and an OpenCLIP text encoder,
while remaining a latent diffusion model
in the same overall design space.\\
\textbf{Architecture variation.} SDXL 
(UNet-2.6B with dual CLIP encoders)~\citep{podell2023sdxl} and 
PixArt-$\alpha$ (DiT transformer with T5-XXL) ~\citep{chen2023pixart}
differ from the UNet-860M family in both 
denoiser architecture and text encoder.\\
\textbf{Training procedure variation.} 
SDXL-Turbo ~\citep{sauer2024adversarial} shares the SDXL architecture but 
is trained via adversarial distillation rather 
than standard diffusion training, isolating 
the effect of training procedure.\\
Together these eight models form a challenging 
attribution setting. 





\textbf{Data.} For each model we generate signatures from $N{=}50$ images and evaluate 
on $N_{\text{test}}{=}500$ held-out images from SD
prompt dataset. We additionally evaluate 
cross-domain generalization by testing the same 
prototypes on MS-COCO prompts \citep{lin2014microsoft} generated images.\\
\textbf{Metrics.} We report Top-1 attribution accuracy for all classifiers across all 
experiments.

\textbf{\textsc{SDS} implementation.} We use the SD\,v1.5 
KL-f8 VAE as the shared encoder ($512{\times}512$ 
input, $64{\times}64$ latents). The probe scheduler 
is DDIM with $T{=}50$ steps. We use $K{=}8$ 
frequency rings, $S{=}3$ amplitude scales 
$\{1.0,\,1.25,\,1.5\}$, $R{=}10$ noise repeats, 
and $T_n{=}5$ evenly spaced timesteps in 
$[0.2T,\,T]$, giving signature dimension 
$S{\cdot}T_n{\cdot}K^2 {=} 960$. LinearSVC is 
trained on raw signatures using 5-fold 
cross-validation. The choice of $K$ and $S$ is studied in the ablation (Appendix \ref{app:ablation}); $R{=}10$ 
and $T_n{=}5$ were selected based on 
convergence of the coupling estimate and 
accuracy saturation respectively.


\noindent{\bf Baselines.} We compare \textsc{SDS} against two non-invasive attribution methods that, 
like \textsc{SDS}, attribute generated images 
to their source model without modifying model 
parameters.
Both are 
evaluated under identical conditions: the same eight 
models, the same test images, and the same evaluation 
protocol as \textsc{SDS}. \textsc{LatentTracer}~\citep{wang2024trace} attributes 
images by inverting the VAE decoder and measuring 
reconstruction fidelity under each candidate model. \textsc{RONAN}~\citep{wang2023did} attributes images 
via gradient-based input inversion with random 
initialization, reverse-engineering the latent input 
of each candidate model.
\vspace{-10pt}

%% file: neurips_2026/arxiv_section/7_results_arxiv.tex
\section{Results}
\label{sec:results}
\vspace{-5pt}
Our evaluation addresses following core questions:
\begin{squishenumerate}
  \item \textbf{Attribution accuracy} 
    (\S\ref{subsec:discriminability}): Can \textsc{SDS} reliably 
    identify the source model across eight diverse 
    diffusion models spanning different architectures, 
    training data, and fine-tuning procedures?
    \item \textbf{Cross-domain generalization} 
    (\S\ref{subsec:generalization}): Do 
    \textsc{SDS} signatures remain stable when 
    query images are drawn from a completely 
    different prompt distribution than the 
    prototypes?
    \item \textbf{Robustness} 
    (\S\ref{subsec:robustness}): Do \textsc{SDS} signatures 
    remain stable when query images are subject to 
    common post-processing distortions?
    \item \textbf{Baseline comparison} 
    (\S\ref{subsec:baseline}): How does \textsc{SDS} perform compared to the non-invasive attribution baseline, on the same 
    eight-model closed-set task?
\end{squishenumerate}

\subsection{Attribution Accuracy}
\label{subsec:discriminability}
\begin{wrapfigure}{r}{0.7\textwidth}
  \centering
  \vspace{\baselineskip}
  \includegraphics[width=1.0\linewidth]{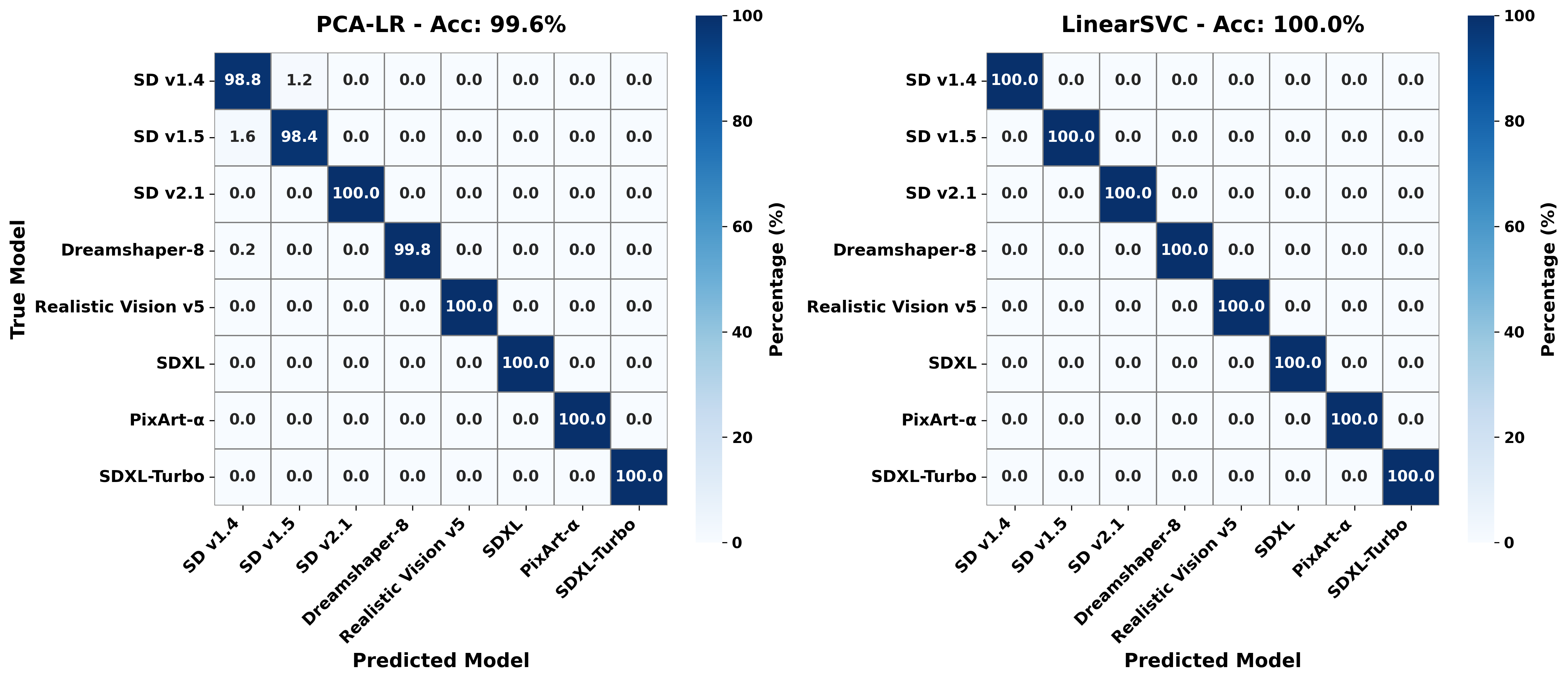}
  \caption{\footnotesize
  \textbf{Confusion matrices for \textsc{SDS} attribution on 
    the SD-Prompts dataset ($M{=}8$ models, 
    $N_{\text{test}}{=}500$ per model).} 
    PCA-LR and LinearSVC 
  at $T_n{=}5$. 
    \vspace{-40pt}}
\label{fig:confusion_1t5t}
\end{wrapfigure}

Figure~\ref{fig:confusion_1t5t} shows confusion 
matrices for PCA-LR and LinearSVC at five 
probe timesteps. Attribution accuracy scales 
directly with model difference: SDXL, 
PixArt-$\alpha$, and SDXL-Turbo are attributed 
perfectly across all settings, as are SD\,v2.1, 
Dreamshaper-8, and Realistic Vision\,v5 despite 
sharing the UNet-860M backbone. The only errors 
occur between SD\,v1.4 and SD\,v1.5 
(significant weight overlap), the hardest 
possible case. LinearSVC achieves 
$\approx {100\%}$ overall accuracy, eliminating 
all confusion between this pair. PCA-LR retains ${\sim}1.5\%$ residual 
confusion even at $T_n{=}5$, consistent with PCA 
compression discarding marginal discriminative 
signal. t-SNE visualizations confirming the geometric 
separability of \textsc{SDS} signatures are 
provided in Appendix~\ref{app:tsne}.
\vspace{-20pt}

\subsection{Prompt Robustness and Cross-Domain 
Generalization}
\label{subsec:generalization}
\begin{wraptable}{r}{0.5\textwidth}
\vspace{-\baselineskip}  
\centering
\scriptsize
\renewcommand{\arraystretch}{1.0}
\setlength{\tabcolsep}{6pt}
\begin{tabular}{l ccc}
\toprule
\textbf{Classifier} 
  & \textbf{SD-Prompts} 
  & \textbf{MS-COCO} 
  & \textbf{Drop (pp)} \\
\midrule
Argmin$^\dagger$     
  & 78.33\% & 78.40\% & $+$0.07 \\
\midrule
LinearSVC           
  & \textbf{99.98\%} & \textbf{96.20\%} & $-$3.80 \\
PCA-LR               
  & 99.62\% & 92.50\% & $-$7.12 \\
Random Forest        
  & 96.23\% & 83.60\% & $-$12.63 \\
PCA-SVM (RBF)        
  & 91.83\% & 80.30\% & $-$11.53 \\
PCA-kNN ($k{=}5$)   
  & 82.37\% & 74.20\% & $-$8.17 \\
\bottomrule
\end{tabular}
\caption{\footnotesize \textbf{Top-1 accuracy 
  across prompt distributions.} Prototypes from 
  SD-Prompts ($N{=}50$, $T_n{=}5$, $M{=}8$, 
  $N_{\text{test}}{=}500$), evaluated unchanged 
  on MS-COCO. Drop: absolute Top-1 decrease. 
  $\dagger$ training-free.\vspace{-4mm}}
\label{tab:main}
\end{wraptable}
We test whether signatures reflect the model 
or the images used to construct them. Prototypes 
are built from SD-Prompts images and evaluated 
unchanged on MS-COCO images, a completely 
different prompt distribution, without any 
update (Table~\ref{tab:main}).

Two patterns emerge. First, the argmin variant 
(Eq.~\ref{eq:argmin}) is essentially unaffected 
($+0.07$pp), confirming that prototype centroids 
are content-invariant by design: row normalisation 
(Eq.~\ref{eq:rownorm}) and unconditional 
conditioning remove content-scale variation so 
completely that the centroid does not shift under 
distribution change. Second, classifier-based 
methods show larger drops, with LinearSVC losing 
$3.78$pp ($99.98\%{\to}96.20\%$) and other 
classifiers dropping by $7$--$13$pp. This reflects 
a different mechanism: classifiers trained on 
SD-Prompts signatures learn decision boundaries 
that are partially distribution-specific, and 
these boundaries shift slightly under MS-COCO. 
LinearSVC is the most robust because its linear 
decision surface exploits the global mean-shift 
structure, which is more stable than the 
nonlinear boundaries learned by Random Forest 
or SVM-RBF. LinearSVC still achieves $96.20\%$ on MS-COCO 
with no retraining; an evidence that the 
underlying signatures are substantially 
model-intrinsic, even if the classifier boundary is not fully distribution-invariant.

\subsection{Robustness to Image Distortions}
\label{subsec:robustness}
\vspace{-5pt}
\begin{wraptable}{r}{0.65\textwidth} 
\vspace{-\baselineskip} 
\centering

\scriptsize
\setlength{\tabcolsep}{2pt}
\begin{tabular}{l rrr rrr}
\toprule
& \multicolumn{3}{c}{\textbf{SD-Prompts}} 
& \multicolumn{3}{c}{\textbf{MS-COCO}} \\
\cmidrule(lr){2-4}\cmidrule(lr){5-7}
\textbf{Distortion} 
  & \textbf{Argmin} 
  & \textbf{PCA-LR} 
  & \textbf{LinearSVC}
  & \textbf{Argmin} 
  & \textbf{PCA-LR} 
  & \textbf{LinearSVC} \\
\midrule
Rotation ($75^\circ$)    
  & 77.5\% & 90.6\% & 91.9\% 
  & 62.5\% & 90.0\% & 90.6\% \\
JPEG (Q=25)              
  & 76.2\% & 91.2\% & 93.1\% 
  & 55.6\% & 86.9\% & 85.0\% \\
Crop (75\%) + resize     
  & 78.1\% & 91.2\% & 90.6\% 
  & 56.3\% & 86.9\% & 87.5\% \\
Brightness ($\times1.5$) 
  & 75.6\% & 88.8\% & 88.8\% 
  & 48.8\% & 86.3\% & 85.6\% \\
\midrule
Noise ($\sigma{=}0.1$)  
  & 61.9\% & 92.5\% & 93.8\% 
  & 58.1\% & 93.8\% & 92.5\% \\
Blur ($\sigma{=}8$)      
  & 51.9\% & 95.6\% & 95.6\% 
  & 21.3\% & 99.4\% & \textbf{99.9\%} \\
\bottomrule
\end{tabular}

\caption{\footnotesize
  \textbf{Robustness of \textsc{SDS} to strong image-level 
  distortions} ($M{=}8$, $N{=}50$ per model, 20 test images per model, $N_{\text{test}}{=}160$ total per distortion). Each distortion is applied to 
  the query image before signature extraction; 
  prototypes are always built from clean images. 
  Distortions are intentionally chosen at challenging 
  magnitudes that substantially degrade perceptual 
  image quality. 
\vspace{-4mm}}
\label{tab:robustness}
\end{wraptable}


We evaluate \textsc{SDS} against six strong post-processing 
distortions applied to query images before signature 
extraction (Table ~\ref{tab:robustness}). These distortions 
are intentionally chosen at challenging magnitudes 
that substantially degrade perceptual image quality; 
prototypes are always built from clean images.

The six distortions split into two behaviours. 
Rotation, JPEG, cropping, and brightness cause 
no meaningful degradation (${\leq}1.3$pp argmin 
drop; LinearSVC ${\geq}88\%$) because they 
preserve latent structure. Noise and blur hurt 
argmin substantially ($61.9\%$ and $51.9\%$), 
but classifier accuracy is \emph{higher} than 
baseline: LinearSVC reaches $93.8\%$ under noise 
and $\mathbf{99.9\%}$ under blur on MS-COCO. This 
is not paradoxical; strong distortions add 
isotropic noise that blurs individual-sample 
distances but leaves class centroids intact, 
and LinearSVC operates on centroids. The 
takeaway is fundamental i.e., \textsc{SDS} measures denoiser 
spectral routing, not pixel statistics, so 
pixel-space distortions cannot erase the 
fingerprint regardless of strength.
Additional ablation results are provided in 
the appendix: sensitivity to prototype set 
size is studied in Appendix~\ref{subsec:prototype}, where we 
justify the choice of $N{=}50$; and 
sensitivity to probe design is studied in 
Appendix~\ref{app:ablation}, where we show 
that attribution saturates at $K{=}6$, 
$S{=}3$ ($99.6\%$), with the baseline 
$K{=}8$, $S{=}3$ operating in this 
saturated regime.

\subsection{Baseline Comparison}
\label{subsec:baseline}
\begin{wraptable}{r}{0.50\textwidth} 
\vspace{-3\baselineskip} 
\centering
\renewcommand{\arraystretch}{1.0}
\small
\setlength{\tabcolsep}{2pt}
\begin{tabular}{l ccc}
\toprule
\textbf{Model} 
  & \textbf{{RONAN}} 
  & \textbf{{LatentTracer}} 
  & \textbf{\textsc{SDS} (ours)} \\
\midrule
SD\,v1.4          & 21.4\% & 22.1\% & \textbf{100.0\%} \\
SD\,v1.5          & 20.2\% & 19.7\% & \textbf{100.0\%} \\
SD\,v2.1          & 22.2\% & 26.6\% & \textbf{100.0\%} \\
Dreamshaper-8     & 21.2\% & 22.1\% & \textbf{99.8\%} \\
Realistic V.\,v5  & 20.4\% & 20.9\% & \textbf{100.0\%} \\
SDXL              & 48.4\% & 59.2\% & \textbf{100.0\%} \\
PixArt-$\alpha$   & 55.4\% & 79.0\% & \textbf{100.0\%} \\
SDXL-Turbo        & 35.0\% & 67.0\% & \textbf{100.0\%} \\
\midrule
\textbf{Overall Top-1} 
  & 30.5\% & 44.8\% & \textbf{99.9\%} \\
\bottomrule
\end{tabular}
\caption{\footnotesize
  \textbf{Baseline comparison on the 8-model 
  closed-set task} ($N_{\text{test}}{=}500$ per model, Top-1 
  accuracy per model). Both baselines exploit 
  decoder reconstruction fidelity and collapse 
  for all five models sharing the SD KL-f8 VAE. 
  \textsc{SDS} probes the denoiser and succeeds across 
  all models.
}
\label{tab:baseline_comparison}
\end{wraptable}


Table~\ref{tab:baseline_comparison} reveals a 
structural gap. Both baselines rely on decoder 
reconstruction as their discriminative signal, 
effective when autoencoders differ 
(\textsc{LatentTracer}: $59$-$79\%$; 
\textsc{RONAN}: $35$-$55\%$ on SDXL, 
PixArt-$\alpha$, SDXL-Turbo), but collapsing 
to $20$-$27\%$ for all five shared-VAE models, 
indistinguishable from the $20\%$ random 
baseline. This is structural: shared decoders 
produce identical reconstruction losses 
regardless of which candidate is queried. 
\textsc{SDS} probes the denoiser instead, 
achieving perfect attribution on all eight models i.e., a $55$pp gain over \textsc{LatentTracer} and 
$70$pp over \textsc{RONAN}, with the entire gap 
on exactly the model pairs that matter most. 

\vspace{-10pt}

%% file: neurips_2026/arxiv_section/9_conclusion_arxiv.tex
\section{Conclusion and Discussion}
\vspace{-5pt}

\textsc{SDS} achieves perfect attribution across 
eight diverse diffusion models, succeeding 
precisely where existing non-invasive baselines 
collapse. Our results reveal that model identity 
is encoded in the denoiser's spectral geometry 
rather than generated outputs: training data, 
architecture, and training procedure each leave 
distinct imprints on frequency-band energy 
routing, grounding attribution in the learned 
score function.\\
Beyond the empirical results, our work raises 
broader questions. First, the mean-shift 
structure, where class centroids are linearly 
separable while individual samples overlap ,
suggests that spectral coupling signatures may 
generalize beyond attribution to other model 
analysis tasks such as detecting fine-tuning, 
quantization artifacts, or distillation. Second, 
the modest cross-domain drop ($3.8$pp for 
LinearSVC under a completely different prompt 
distribution) raises the question of what would 
be required to deliberately evade \textsc{SDS}: 
an adversary would need to modify the latent 
representation of an image to mimic a different 
model's denoising geometry, a substantially 
harder task than pixel-level post-processing. 
Third, the collapse of decoder-based baselines 
points to a general principle: attribution 
signal should be sought in the component that 
varies most across models i.e., the denoiser, not 
the decoder.\\
\textsc{SDS} is a provenance tool; it answers 
which model produced an image, not whether 
the image has been modified after generation, 
and should be combined with tamper detection 
when image integrity cannot be assumed. 
Natural extensions include open-set attribution, 
where the source model may not appear in the 
registry, and scaling prototype construction 
to larger model registries.
\vspace{-10pt}

\section{Limitations and Broader Impact}
\label{sec:limitations}
\vspace{-5pt}

\textbf{Limitations.}
\textsc{SDS} has two limitations worth noting. 
First, it operates in a closed-set setting where 
the source model must appear in the registry; 
open-set attribution, where the query image may 
originate from an unknown model, remains an open 
problem. Second, it requires white-box access to 
candidate model weights to run denoiser forward 
passes; this is satisfied by all open-source 
diffusion models but precludes attribution against 
proprietary APIs where weights are unavailable. Finally, \textsc{SDS} does not detect post-generation tampering; combining it with 
tamper detection is an important direction for 
future work.\\
\textbf{Broader impact.}
\textsc{SDS} addresses a genuine need in 
responsible AI deployment: verifying the 
provenance of generated images for ownership 
verification, misuse detection, and content 
moderation. By enabling attribution without 
model modification or watermarking, it is 
applicable to the large ecosystem of existing 
open-source diffusion models that cannot be 
retroactively enrolled in watermarking schemes. 
Attribution tools of this kind are more likely 
to support accountability than to enable misuse. 
One potential concern is that knowledge of 
\textsc{SDS}'s mechanism could inform 
adversarial attacks designed to spoof or evade 
attribution; studying such attacks and defenses 
is an important direction for future work.

%% file: neurips_2026/arxiv_section/appendix_arxiv.tex
\appendix
{\large \begin{center} {\bf APPENDIX} \end{center}}

\section{Algorithm}
\label{app:algorithm}

\input{neurips_2026/figures/algorithm}

\section{Extended Related Work}
\label{app:related}

\noindent{\bf Watermarking for diffusion models.}
A dominant line of work on protecting diffusion model intellectual property is \emph{watermarking}, which embeds identifiable signals into model parameters or generated outputs. Existing approaches introduce watermarks through training-time modifications, such as embedding triggers in the data~\citep{zhao2023recipe, peng2025intellectual}, fine-tuning model components~\citep{yuan2024watermarking}, or altering the decoder to encode persistent signatures in generated images~\citep{fernandez2023stable}. While effective, these methods are inherently \emph{invasive}, requiring retraining or fine-tuning and potentially affecting generation quality or model usability. Moreover, watermarks can be removed or degraded through post-processing or model modifications~\citep{hu2024stable}, motivating the need for non-invasive alternatives.

\noindent{\bf Model fingerprinting in neural networks.}
Non-invasive fingerprinting aims to identify intrinsic characteristics of a model without modifying its parameters. In classification models, prior work exploits decision boundary geometry or adversarial examples to construct model-specific signatures~\citep{cao2021ipguard, lukas2019deep, peng2022fingerprinting}. In image restoration, fingerprinting methods identify \emph{critical points} near the performance boundary that produce distinctive responses across models~\citep{quan2023fingerprinting}. These approaches rely on the observation that models differ in how they behave on carefully chosen inputs. However, extending these ideas to diffusion models is non-trivial, as diffusion models are probabilistic generative systems that map noise distributions to images rather than deterministic input-output mappings.

\section{Theoretical Grounding}
\label{subsec:theory}

\noindent{\bf Score function and spectral geometry.}
The denoising network $\hat{\epsilon}_\theta(z_t, t)$ 
is related to the score function via 
$\nabla_{z_t} \log p_t(z_t) \propto 
-\,\hat{\epsilon}_\theta(z_t, t)$~\citep{song2020score}. 
The score field $\nabla_{z_t} \log p_t(z_t)$ 
encodes the statistical geometry of the 
(noisy) data distribution at noise level $t$,
including how energy is correlated across spatial 
frequency bands. Concretely, the Hessian 
$\nabla^2_{z_t}\log p_t(z_t)$ characterizes the local 
covariance structure: its off-diagonal blocks in the 
frequency domain reflect how variations in one 
frequency band couple to variations in another.
Different training datasets imprint different covariance 
structures, and different architectures impose 
different inductive biases on how this structure 
is represented~\citep{kadkhodaie2023generalization}. 
Both effects manifest in the Jacobian of the 
denoiser $J_\theta(z_t,t) = \partial\hat{\epsilon}_\theta(z_t,t) 
/ \partial z_t$, which inherits the spectral 
geometry of the learned score function.

\noindent{\bf Spectral coupling as a Jacobian 
projection.}
The coupling matrix $C[k,j]$ is a Monte Carlo estimate 
of this Jacobian structure in the frequency domain.
To see this, consider a first-order expansion of the 
denoiser response to a band-limited perturbation 
$\eta^{(k,s)}$ confined to ring $k$:
\begin{equation}
  \hat{\epsilon}_\theta(z_t, t, \varnothing) 
  \approx 
  \hat{\epsilon}_\theta(z_0, t, \varnothing) 
  + J_\theta(z_0, t)\,
  \sqrt{1{-}\bar{\alpha}_t}\,\eta^{(k,s)},
  \label{eq:taylor}
\end{equation}
where $z_t = z_0 + \sqrt{1{-}\bar{\alpha}_t}\,\eta^{(k,s)}$.
The output residual 
$\delta\hat{\eta} = \hat{\epsilon}_\theta(z_t,t,\varnothing) 
- \hat{\epsilon}_\theta(z_0,t,\varnothing)$
therefore reflects how $J_\theta$ maps ring-$k$ 
inputs to outputs across all rings. The coupling 
entry $C[k,j]$ measures the energy of this 
residual in ring $j$, averaged over noise samples.
If $\eta^{(k,s)}$ is isotropic within ring $k$, then
\begin{equation}
  C[k,j] \;\propto\; 
  \bigl\|P_j \, J_\theta(z_0,t) \, 
  P_k\bigr\|_F^2,
  \label{eq:jacobian_proj}
\end{equation}
where $P_k$ denotes projection onto ring $k$.
Since $J_\theta$ is determined by the learned 
score function, the expected couplings $C[k,j]$ are 
intrinsic to the model and, after averaging over 
images, largely independent of image content.

\section{Signature Separability Across Models}
\label{app:tsne}
\begin{figure}[h]
    \centering
    \includegraphics[width=1\linewidth]{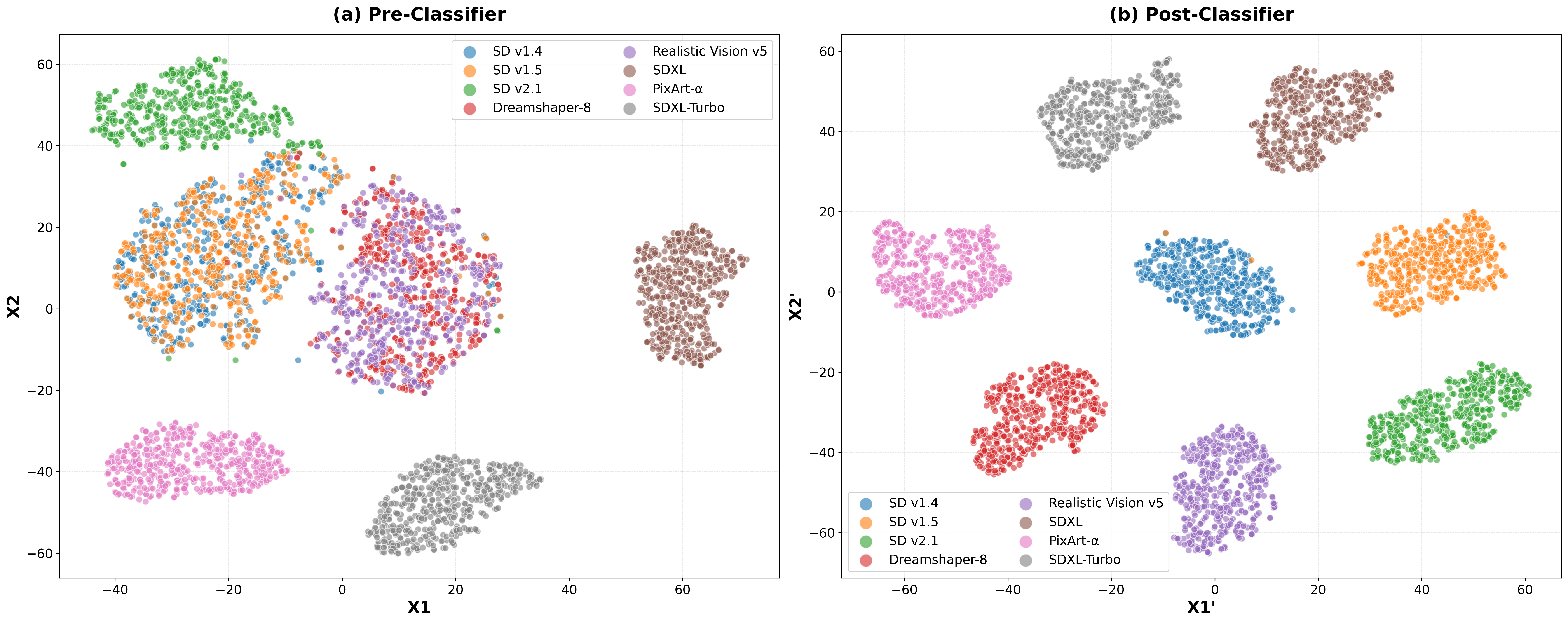}
    \caption{\footnotesize \textbf{t-SNE visualization of SDS signatures 
    before and after classifier projection.} 
    (a)~\textit{Pre-classifier}: t-SNE of dimensionally reduced SDS 
    signatures, with no classifier applied. Four of the eight models 
    form visually distinct clusters, confirming that the signature 
    space carries strong intrinsic structure without any learned 
    transformation. The only overlap occurs between SD\,v1.4 - 
    SD\,v1.5, and Dreamshaper-8 - Realistic Vision v5, 
    which are the hardest pairs in our evaluation.
    (b)~\textit{Post-classifier}: t-SNE of signatures after LR 
    projection. All eight clusters are cleanly separated, consistent with the high attribution 
    accuracy reported in confusion matrix ~\ref{fig:confusion_1t5t}.}
    \label{fig:tsne}
\end{figure}

The t-SNE visualization in Figure~\ref{fig:tsne} provides direct 
geometric evidence for the quality of SDS signatures. Even before 
any classifier is applied, the raw signature space already separates 
most of the models into distinct, compact clusters. This 
intrinsic separability confirms that the spectral coupling 
representation captures genuine model-specific structure rather than 
fitting to classifier artifacts. A linear classifier resolves this 
residual ambiguity entirely, as shown in panel~(b), consistent with 
the global mean-shift structure established in 
\S~\ref{sec:motivation}.

\section{Sensitivity to Probe Design}
\label{app:ablation}

\begin{table}[h]
\centering
\renewcommand{\arraystretch}{1.2}
\small
\setlength{\tabcolsep}{6pt}
\begin{tabular}{cc r r r}
\toprule
\textbf{K} & \textbf{S} & \textbf{Sig.\ dim}& \textbf{Argmin} & \textbf{LinearSVC} \\
\midrule
2 & 1 & 4   & 52.3\% & 52.4\% \\
  & 2 & 8   & 53.5\% & 65.3\% \\
  & 3 & 12  & 60.9\% & 77.4\% \\
\midrule
4 & 1 & 16  & 65.1\% & 90.2\% \\
  & 2 & 32  & 66.8\% & 94.6\% \\
  & 3 & 48  & 70.3\% & 97.2\% \\
\midrule
6 & 1 & 36  & 69.8\% & 94.3\% \\
  & 2 & 72  & 72.4\% & 97.0\% \\
  & 3 & 108 & 75.0\% & 99.6\% \\
\midrule
8 & 1 & 64  & 71.9\% & 94.2\% \\
  & 2 & 128 & 74.2\% & 97.5\% \\
  & 3$^\ast$ & 192 & 76.4\% & \textbf{99.9\%} \\
\bottomrule
\end{tabular}
\caption{ \footnotesize
  \textbf{Ablation over frequency rings $K$ and 
  perturbation amplitudes $S$} at a single probe 
  timestep ($T_n{=}1$, $t=25$), evaluated 
  on all 8 models ($N_{\text{test}}{=}500$ per model). 
  $S{=}1$: amplitude $\{1.0\}$; $S{=}2$: 
  $\{1.0,\,1.25\}$; $S{=}3$: $\{1.0,\,1.25,\,1.5\}$ 
  (full baseline$^\ast$). Sig.\ dim $= S \times K^2$. 
  $^\ast$~baseline used in main experiments.
}
\label{tab:ablation_full}
\end{table}

Table~\ref{tab:ablation_full} reports attribution 
accuracy across all combinations of $K$ and $S$, 
evaluated at $T_n{=}1$ to isolate the effect of 
spatial resolution and multi-scale probing from 
temporal aggregation. Both parameters independently and substantially affect performance. At $K{=}2$, even the full 
$S{=}3$ configuration reaches only $77.4\%$ with 
LinearSVC, near chance for an 8-way task. 
Increasing $K$ from 2 to 4 with $S{=}3$ jumps 
performance to $97.2\%$, confirming that spectral 
resolution is critical: coarser frequency 
partitions cannot distinguish the subtle cross-band 
routing differences that encode model identity. 
Performance saturates beyond $K{=}6$, with 
$K{=}6$, $S{=}3$ already achieving $99.6\%$. The 
baseline configuration ($K{=}8$, $S{=}3$) operates 
in this saturated regime, offering a favorable 
trade-off between performance and signature 
dimensionality.

The effect of $S$ is consistent across all values 
of $K$: adding perturbation amplitudes improves 
performance by $10$--$25$\,pp at fixed $K$, with 
diminishing returns from $S{=}2$ to $S{=}3$, 
suggesting three amplitudes is a practical optimum. 
Different amplitudes probe the denoiser at different 
operating points, each revealing complementary 
model-specific structure.

\section{Sensitivity to Prototype Set Size}
\label{subsec:prototype}

\begin{wrapfigure}{r}{0.55\textwidth}
\vspace{-3\baselineskip}
\centering
\includegraphics[width=0.95\linewidth]
  {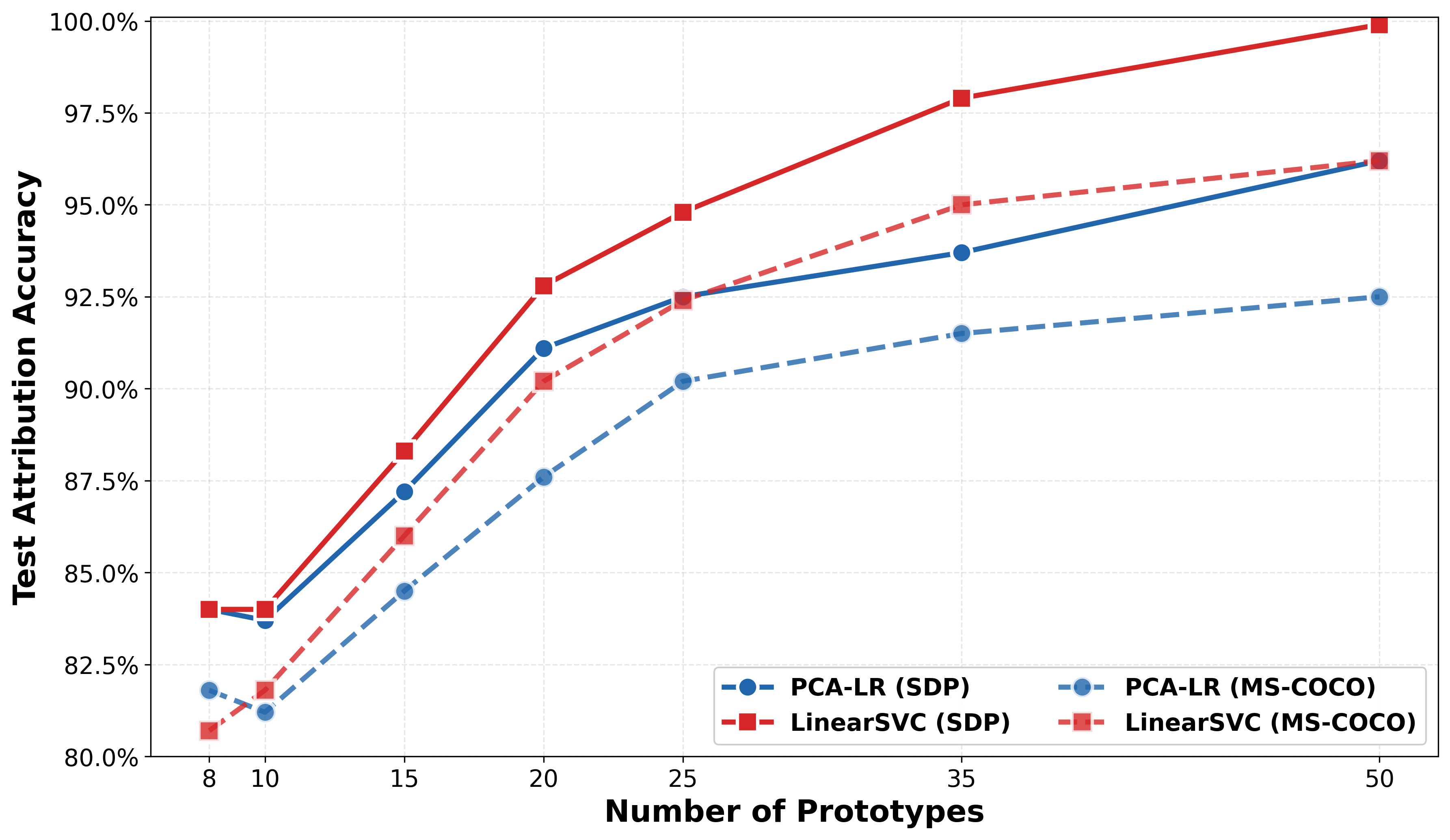}
\caption{\footnotesize \textbf{Attribution 
  accuracy vs.\ number of prototype images 
  per model ($N$).} LinearSVC and PCA-LR 
  evaluated on SD-Prompts (in-distribution) 
  and MS-COCO (cross-domain). \vspace{-4mm}}
\label{fig:prototype_comparison}
\end{wrapfigure}

Figure~\ref{fig:prototype_comparison} shows 
how attribution accuracy scales with $N$, the 
number of prototype images per model used to 
train the classifier. At $N{=}8$ per model 
($64$ total training signatures), LinearSVC 
achieves $84.0\%$ on SD-Prompts and $80.7\%$ 
on MS-COCO, already well above the $12.5\%$ 
random baseline. Accuracy improves steadily 
with $N$: at $N{=}25$, LinearSVC reaches 
$94.8\%$ on SD-Prompts and $92.4\%$ on 
MS-COCO, and at $N{=}50$ it reaches $99.9\%$ 
and $96.2\%$ respectively. The gap between 
SD-Prompts and MS-COCO remains stable across 
all values of $N$ (${\approx}3$--$4$pp), 
confirming that the cross-domain drop reflects 
a fixed distribution shift rather than 
amplified estimation error. PCA-LR follows 
the same trend, consistently $2$--$3$pp below 
LinearSVC. These results confirm that $N{=}50$ 
prototypes per model is a practical and 
sufficient operating point for reliable 
attribution.

%% file: neurips_2026/figures/algorithm.tex
\begin{algorithm}[!ht]
\label{alg:sgs}
\small
\KwIn{Denoiser $\hat{\epsilon}_\theta$, latent $z_0$,
      rings $K$, timesteps $\mathcal{T}$, scales $\mathcal{S}$,
      repeats $R$}
\KwOut{Signature $\Phi \in \mathbb{R}^{S \cdot T_n \cdot K^2}$}

Compute ring masks $\{M_k\}_{k=0}^{K-1}$\;
$\mathbf{A} \leftarrow \mathbf{0}^{S \times T_n \times K \times K}$\;

\For{each repeat $r = 1,\ldots,R$}{
  \For{each $(s,\, t) \in \mathcal{S} \times \mathcal{T}$}{
    \For{each ring $k = 0,\ldots,K-1$}{
      $\eta^{(k,s)} \leftarrow s \cdot \mathcal{F}^{-1}(\mathcal{F}(\eta)\odot M_k)$ \tcp*[r]{band-limited perturbation}

$z_t \leftarrow \sqrt{\bar{\alpha}_t}\,z_0 + \sqrt{1-\bar{\alpha}_t}\,\eta^{(k,s)}$ \tcp*[r]{inject into diffusion}

$\hat{\eta} \leftarrow \hat{\epsilon}_\theta(z_t, t, \varnothing)$ \tcp*[r]{denoiser forward pass}

$r \leftarrow \hat{\eta} - \eta^{(k,s)}$ \tcp*[r]{residual = model response}

$\mathbf{A}[s,t,k,\cdot] \mathrel{+}= \langle |\mathcal{F}(r)|^2, M_j \rangle_{j=0}^{K-1}$ \tcp*[r]{energy redistribution}
    }
  }
}

$\mathbf{A} \leftarrow \mathbf{A} / R$\;\quad
Row-normalise: $\mathbf{A}[s,t,k,\cdot] \leftarrow
   \mathbf{A}[s,t,k,\cdot]\,/\,\|\mathbf{A}[s,t,k,\cdot]\|_1$\;
\Return{$\mathrm{vec}(\mathbf{A})$}\;
\caption{\footnotesize \textbf{Spectral Geometry Signature (SGS) extraction.} The algorithm probes a diffusion model by injecting band-limited perturbations into latent space and measuring how the denoiser redistributes energy across frequency bands over timesteps and scales, producing a model-specific signature  
\vspace{-8pt}}
\end{algorithm}